\documentclass[11pt]{article}

\usepackage[preprint]{acl}

\usepackage{times}
\usepackage{latexsym}
\usepackage{booktabs}
\usepackage{float}
\usepackage[T1]{fontenc}

\usepackage[utf8]{inputenc}

\usepackage{microtype}
\usepackage[most]{tcolorbox}

\usepackage{inconsolata}

\usepackage{graphicx}

\title{Flowchart2Mermaid: A Vision-Language Model Powered System for Converting Flowcharts into Editable Diagram Code}

\author{Pritam Deka \\
  Queen's University Belfast / UK \\
  \texttt{p.deka@qub.ac.uk} \\\And
  Barry Devereux \\
  Queen's University Belfast / UK \\
  \texttt{B.Devereux@qub.ac.uk} \\}

\begin{document}
\maketitle
\begin{abstract}
Flowcharts are common tools for communicating processes but are often shared as static images that cannot be easily edited or reused. We present \textsc{Flowchart2Mermaid}, a lightweight web system that converts flowchart images into editable Mermaid.js code which is a markup language for visual workflows, using a detailed system prompt and vision-language models. The interface supports mixed-initiative refinement through inline text editing, drag-and-drop node insertion, and natural-language commands interpreted by an integrated AI assistant. Unlike prior image-to-diagram tools, our approach produces a structured, version-controllable textual representation that remains synchronized with the rendered diagram. We further introduce evaluation metrics to assess structural accuracy, flow correctness, syntax validity, and completeness across multiple models.
\end{abstract}

\section{Introduction}


Flowcharts are central to communicating processes and structures in software engineering, algorithm design, and education \cite{ensmenger2016multiple,smetsers2017problem}. Yet most flowcharts exist as static images, which limits reuse: changing a label or inserting a decision often requires redrawing the entire figure. Automatically converting flowchart images into structured textual code also paves the way for more flexible multimodal reasoning \cite{suri2025follow,soman2025graph,ye2025beyond}. Flowcharts and related procedural diagrams encode stepwise logic and control flow that are difficult to capture through conventional OCR or shape-recognition pipelines alone. With recent advances in vision language models (VLMs), such as GPT-4.1\footnote{\url{https://openai.com/index/gpt-4-1/}} \cite{achiam2023gpt} and Gemini-2.5 \cite{comanici2025gemini}, it has become possible to infer not only visual structures but also their logical interrelations, enabling systems that translate complex diagram images into expressive, editable representations.

\begin{figure}[h]
  \includegraphics[width=\columnwidth]{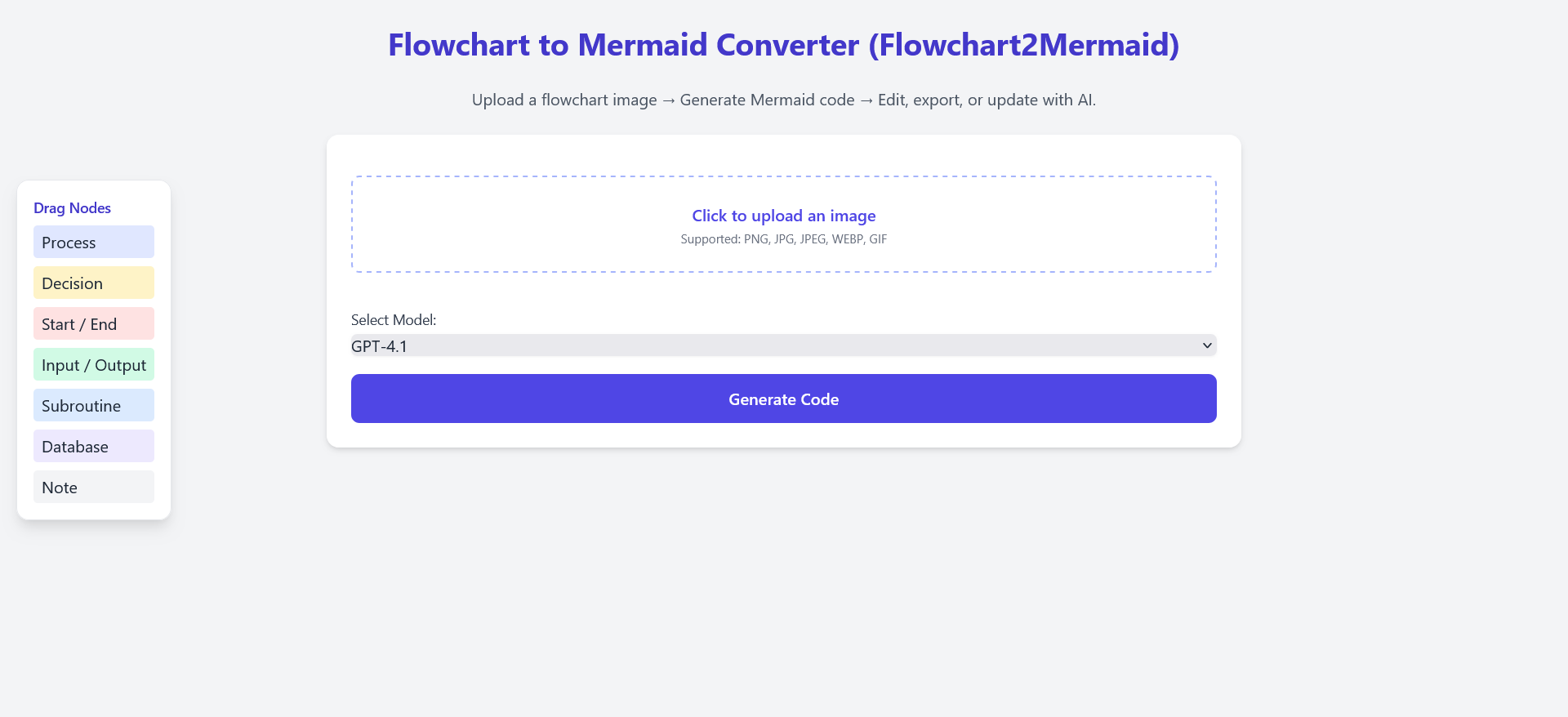}
  \caption{Flowchart2Mermaid Homepage}
  \label{fig:homepage1}
\end{figure}

We introduce \textsc{Flowchart2Mermaid}\footnote{\url{https://flowchart-to-mermaid.vercel.app/}}
, an end-to-end web application that converts flowchart images into textual \texttt{Mermaid.js}\footnote{\url{https://mermaid.js.org/}}
 \cite{sveidqvist2021official} specifications and enables rich human-in-the-loop editing. After uploading an image, a vision–language model (VLM) produces an initial Mermaid program, which users can refine through three complementary interaction modes:
(1) inline editing, by clicking labels directly on the rendered diagram;
(2) visual editing, via drag-and-drop insertion of standard flowchart symbols that automatically synchronize with the underlying code; and
(3) natural-language edits, where free-form instructions trigger targeted LLM modifications to the diagram (\emph{e.g., “Connect \texttt{A} to \texttt{B} with a ‘Yes’ edge”}).
The system supports real-time rendering, export to SVG/MMD formats, and seamless handoff to the Mermaid Live Editor. The landing page of the application is shown in Figure \ref{fig:homepage1}.

Our contributions are: (i) a lightweight, production-ready demo for multimodal diagram-to-code conversion; (ii) a unified interaction design combining textual, visual, and natural-language editing; and (iii) a model-agnostic backend that accommodates multiple VLM providers while maintaining output validity. 
A demonstration video of the work is available for viewing via the Dropbox link\footnote{\url{https://www.dropbox.com/scl/fi/wzoidww81ccqvw2pmwxn8/flowchart2mermaid_demo_pritam.mkv?rlkey=li9d3jliwvqf40kvjdm9r7tgk&st=651lnjfa&dl=0}.}.

\section{Related Work}

This section reviews research relevant to transforming visual diagrams into structured, editable representations. We summarize prior work on diagram parsing, image-to-code translation, and interactive tools for programmatic diagram editing.

\paragraph{Diagram Understanding and Diagram-to-Code.}
Early systems for flowchart interpretation focused on rule-based shape detection and template matching, enabling conversion of hand-drawn or scanned flowcharts into pseudocode or C programs \cite{wu2011research, herrera2017flow2code, carton2013fusion}. Educational environments such as \textit{Flowgorithm} \cite{flowgorithm} or browser-based structured editors \cite{supaartagorn2017web} support generating executable code from diagrams, but require users to construct diagrams within the tool itself and cannot process arbitrary images. More recent approaches incorporate deep-learning pipelines: Arrow-RCNN \cite{schafer2019arrow} and DrawnNet \cite{fang2022drawnnet} use CNNs to detect shapes and connectors, while Montellano et~al.\ \cite{montellano2022recognition} reconstruct both the visual diagram and corresponding code from hand-drawn sketches\footnote{\url{https://github.com/dbetm/handwritten-flowchart-with-cnn}}. Although effective within controlled datasets, these systems lack generalization to diverse diagram styles and do not support interactive refinement.

\paragraph{Multimodal Models for Structured Visual Content.}
Advances in vision language models (VLMs) have enabled more flexible diagram parsing. The \textit{GenFlowchart} system \cite{arbaz2024genflowchart} combines SAM-based (Segment Anything Model) \cite{kirillov2023segment} segmentation with an LLM to assemble a unified symbolic representation from detected visual elements. Foundation models such as GPT-4 \cite{achiam2023gpt} and Gemini \cite{team2023gemini} can translate images into structured text or code, demonstrating strong priors about graphical layout and syntax. Recent work has extended these capabilities to \emph{business process diagrams}  \cite{deka2025structuredextractionbusinessprocess} where the authors present a VLM-based pipeline for extracting structured JSON representations from BPMN images, enriched with OCR-based label recovery and alignment against XML ground truth. These multimodal systems enable diagram-to-text generation as a high-level translation task, moving beyond handcrafted rules. However, prior work typically treats generation as a one-shot inference problem and does not integrate downstream editing workflows.

\paragraph{Interactive Editing and Natural Language-driven Diagram Manipulation.}
A growing ecosystem of AI-assisted diagram editors has recently emerged, including commercial platforms such as FlowchartAI\footnote{\url{https://flowchartai.com}}, dAIgram\footnote{\url{https://www.daigram.app/?utm_source=eliteai.tools&ref=eliteai.tools}}, which convert uploaded images into editable flowcharts using shape detection and layout reconstruction. Tools such as DiagramGPT\footnote{\url{https://www.eraser.io/diagramgpt}} and Lucidchart AI\footnote{\url{https://www.lucidchart.com/pages/use-cases/diagram-with-AI}} provide natural-language (NL) driven diagram creation, while yEd Live\footnote{\url{https://www.yworks.com/products/yed-live}} integrates ChatGPT for modifying existing digital graphs (e.g., renaming or styling nodes). 
Although these systems demonstrate promising forms of AI-assisted editing, they cover only parts of the broader diagram–understanding space. 
As a result, existing tools rarely support semantic editing, version-control-friendly outputs, or deep coupling between recognition, code generation, and iterative refinement. They further lack bidirectional synchronization between visual edits and a symbolic diagram representation, and their natural-language capabilities are typically limited to high-level prompting rather than fine-grained, code-level refinement.

In contrast to the systems above, our approach targets the full pipeline from \emph{raw diagram images} to \emph{textual}, semantically meaningful \texttt{Mermaid} code, supporting downstream reasoning and reproducible version control. While recent work \cite{deka2025structuredextractionbusinessprocess} demonstrates robust VLM-based extraction for BPMN, their focus is on faithful structured recovery rather than interactive editing. Beyond initial multimodal parsing, our interface enables iterative refinement through three coordinated interaction modes: inline editing of text within the rendered diagram, drag-and-drop insertion of common flowchart elements, and natural-language commands that trigger targeted code modifications. All edits such as visual, textual, or linguistic are synchronized with the underlying \texttt{Mermaid} program, maintaining a consistent representation throughout. To our knowledge, no prior system combines diagram-image understanding with a bi-directional code–visual editor and natural-language manipulation, making our work distinct within both AI-assisted diagram editing and multimodal understanding research.

\section{System Overview}

\begin{figure}[t]
  \includegraphics[width=\columnwidth]{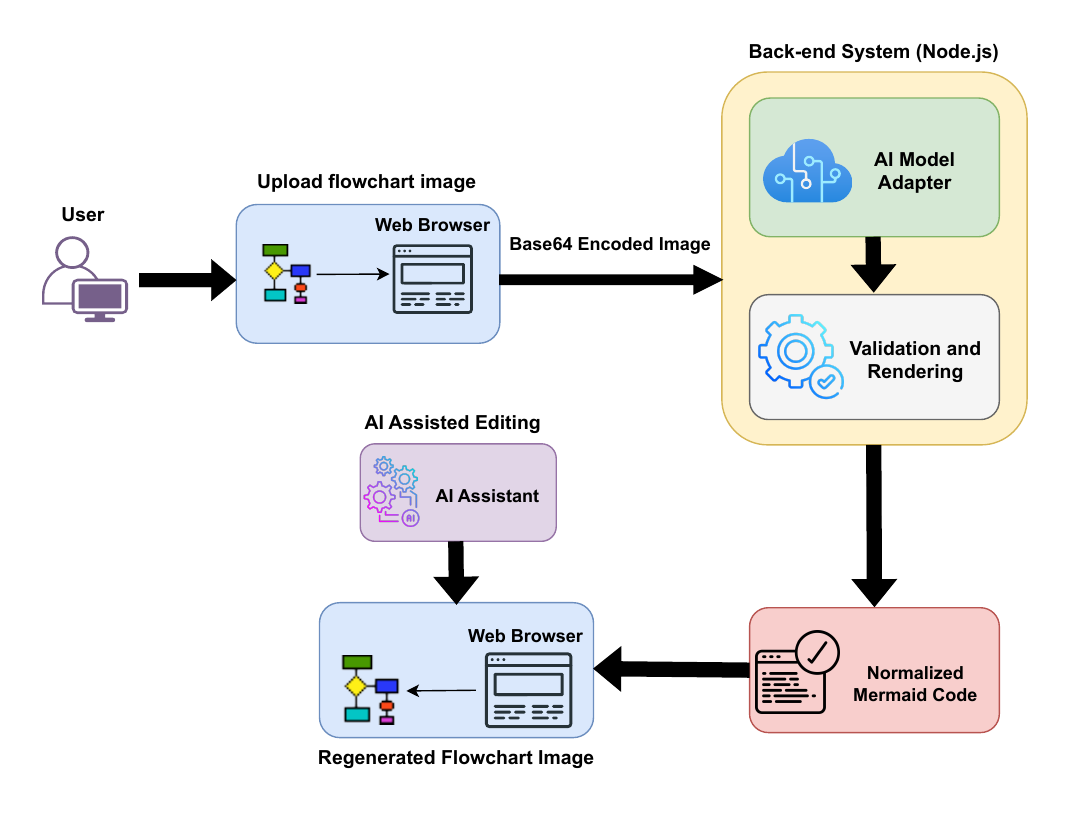}
  \caption{System Overview}
  \label{fig:architecture}
\end{figure}

Figure~\ref{fig:architecture} provides a high-level view of the 
\textsc{Flowchart2Mermaid} system. The application follows a lightweight client–server design in which the browser is responsible for visualization and user interaction, while a minimal server component mediates communication with external vision–language models (VLMs).

At a conceptual level, the system supports a mixed-initiative workflow: a user uploads a flowchart image, obtains an initial \texttt{Mermaid} representation generated by a multimodal model, and then iteratively refines it through direct manipulation, text editing, or natural-language commands. Throughout the workflow, the textual and visual representations remain tightly synchronized, supporting seamless switching between code-level and diagram-level editing.

\section{Implementation}

The system is implemented as a lightweight web application integrating VLM inference with an interactive visualization layer. Its architecture balances modularity, responsiveness, and reproducibility across multiple model back ends.

\subsection{Architecture Overview}

The system is composed of a browser-based front end for visualization and editing and a lightweight Node.js back end that handles multimodal model inference. This separation keeps user interactions responsive while delegating computationally intensive processing to the server.

\begin{figure}[h]
  \includegraphics[width=\columnwidth]{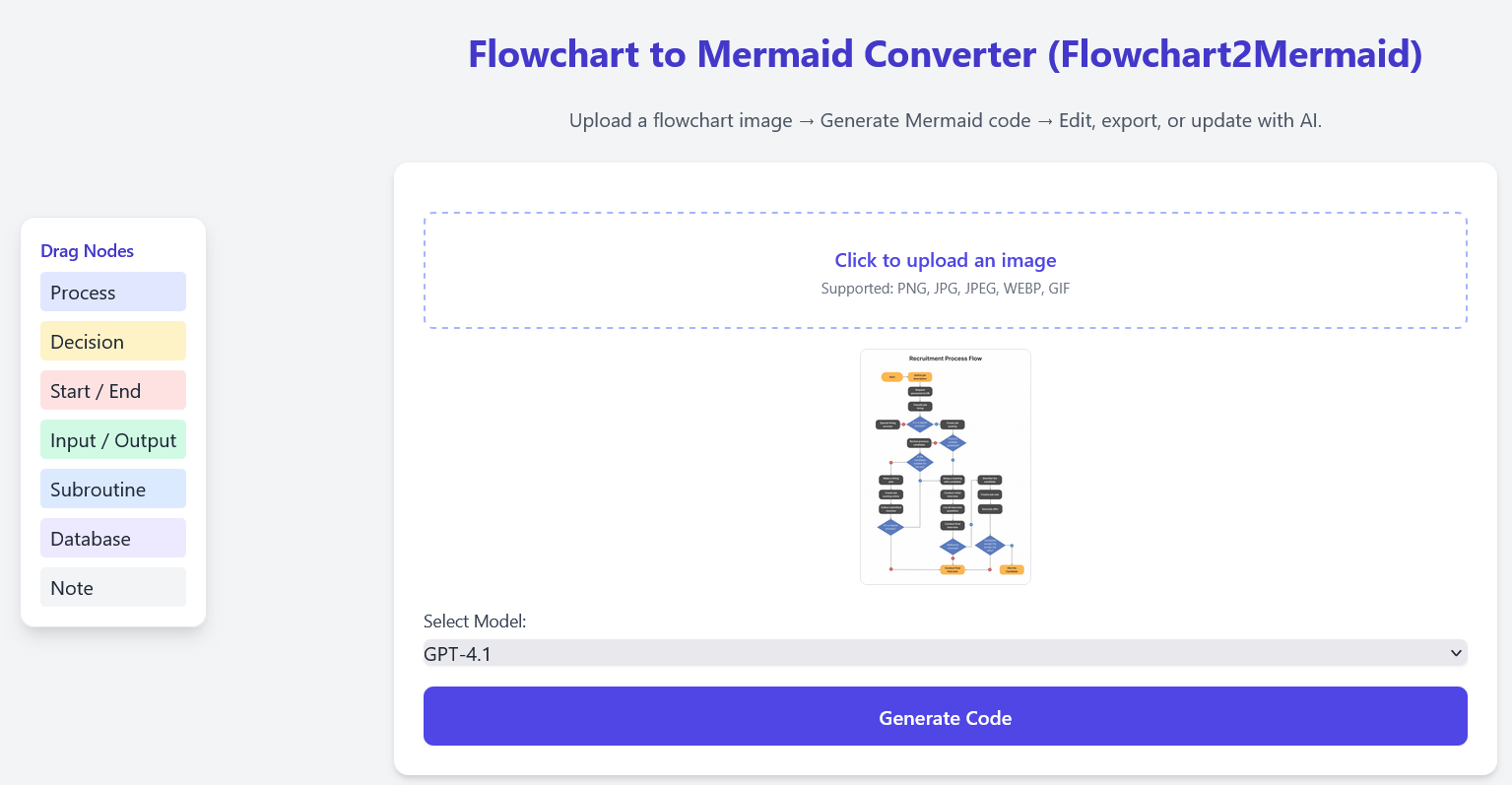}
  \caption{Uploaded example flowchart image}
  \label{fig:fullshot1}
\end{figure}

\begin{figure}[h]
  \includegraphics[width=\columnwidth]{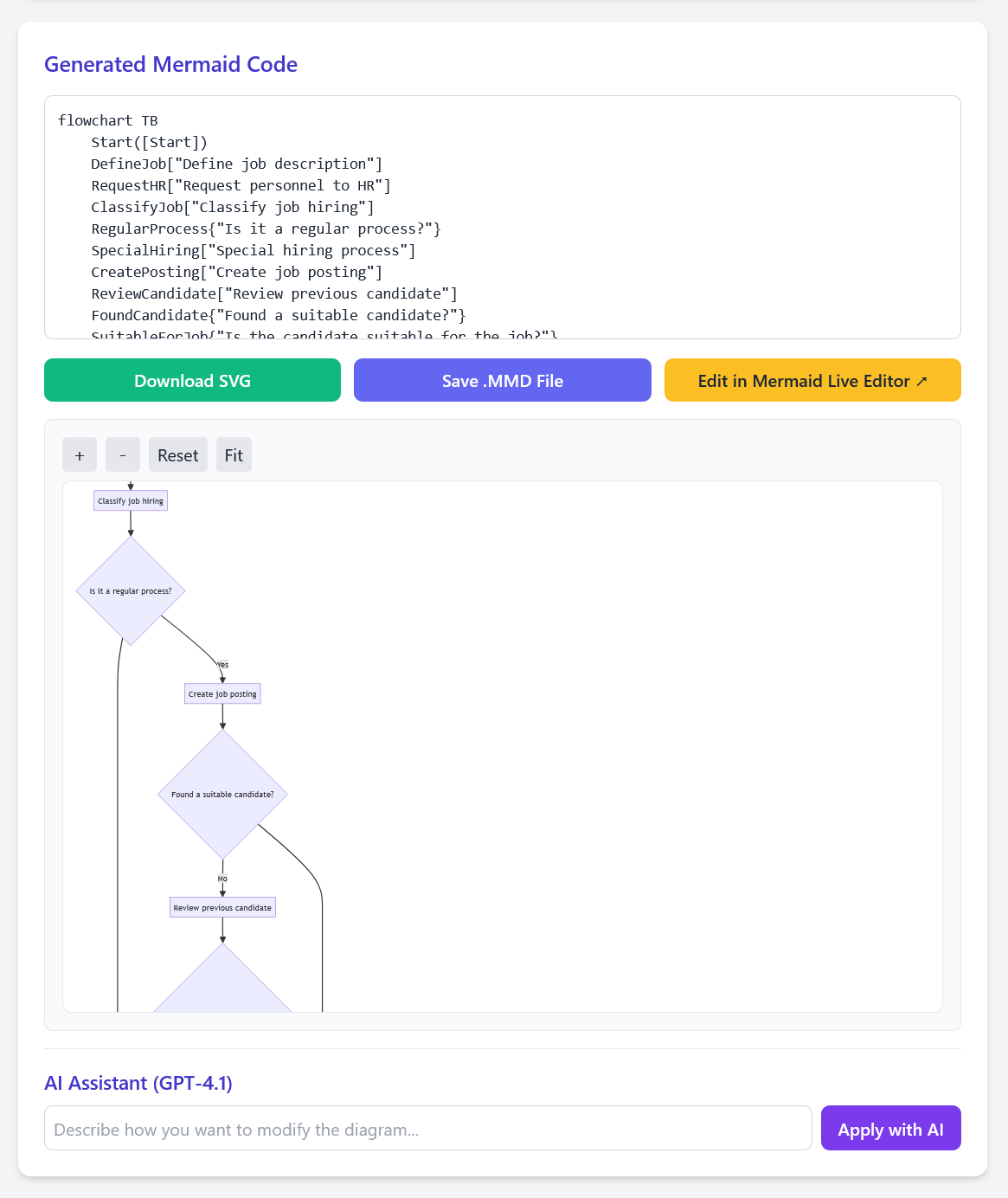}
  \caption{Generated mermaid code from example flowchart image}
  \label{fig:fullshot2}
\end{figure}

After a user uploads a diagram image as shown in Figure \ref{fig:fullshot1}, the front end encodes it and forwards it, together with the chosen model identifier, to the back end. Inference is guided by a carefully engineered system prompt that specifies the expected flowchart elements, structural constraints, and output format of valid \texttt{Mermaid} code. This prompt serves as the core of the conversion process, drawing on insights from structured prompting and constrained generation \cite{liu2023pre, cheng2024structure}. The VLM produces a first-pass \texttt{Mermaid} program, which is lightly normalized before being returned to the browser as shown in Figure \ref{fig:fullshot2}.

On the client side, the code is rendered immediately and remains fully editable. Users may refine the diagram through direct manipulation (dragging nodes, adding connections, modifying labels), through text-based editing of the \texttt{Mermaid} program, or via natural-language commands handled by an integrated AI assistant consisting of the GPT-4.1 model. All edits maintain bidirectional synchronization between the textual and visual representations. This architecture supports rapid iteration while remaining model-agnostic and easily extensible. We showcase the complete working demo for the example flowchart image in video.

\subsection{Front-end Interface}
The front end is implemented using \texttt{HTML}, \texttt{TailwindCSS}, and \texttt{JavaScript}, and provides an interactive diagram-editing environment powered by \texttt{Mermaid.js}. Generated flowcharts are rendered inside a dedicated scrollable viewport that supports controlled zoom (via UI buttons) and smooth navigation, ensuring that large diagrams remain viewable without requiring excessive page scrolling. All elements are fully responsive, and the visual canvas automatically scales diagrams to fit the available space.

Users can modify diagrams directly through inline editing of node labels, drag-and-drop insertion of common flowchart primitives (e.g., processes, decisions, I/O nodes, and databases), and structural operations such as repositioning or deleting elements. The system supports exporting diagrams as \texttt{.mmd} or \texttt{.svg} files and provides one-click access to the Mermaid Live Editor for advanced editing or collaboration.

An embedded \textbf{AI assistant} allows natural-language modification of diagrams. For example, users can issue commands like ``\textit{add a decision node before process C}'' or ``\textit{connect start to review}'', which are parsed into structured transformations of the current Mermaid graph. The interface maintains bidirectional consistency between textual and graphical representations, ensuring that each update is immediately reflected in both views.

\subsection{Back-end Functionality}
The back end orchestrates the conversion and refinement process. Two primary operations are supported:

\noindent
\textbf{Image-to-code generation:} The back end receives an encoded diagram image and uses the chosen model to generate Mermaid syntax. The output is cleaned of Markdown artifacts, normalized for consistent directionality, and validated.\\
\textbf{Code refinement:} The current code and user instruction are combined into a controlled prompt, enforcing deterministic decoding parameters for reproducibility. The resulting update is validated and returned for rendering.

This unified design abstracts away API-specific differences between providers, offering a consistent interface to the front end while enabling easy extension to future multimodal models.

\subsection{System Reliability and Performance}
Strict validation routines ensure that all inputs conform to supported image types and that generated outputs compile successfully in \texttt{Mermaid.js}. 
The architecture emphasizes modular extensibility and low-latency feedback, making it suitable for both research exploration and pedagogical use cases involving diagrammatic reasoning.

\section{Evaluation}

We evaluate our flowchart-to-Mermaid pipeline using a subset of the \textsc{FlowVQA} dataset \cite{singh2024flowvqa}, which contains diverse, real-world flowcharts annotated with step-level semantics. Although intended for visual question answering, its Mermaid annotations for each image allow us to construct the gold-standard (image, code) pairs for our task.

Because Mermaid code admits multiple semantically equivalent surface forms, direct string matching is unreliable: models may emit different node identifiers, paraphrased labels, or alternate arrow styles while still producing a fully correct diagram. These variations make rigid token- or rule-based evaluation brittle. Following recent work on \emph{LLM-as-a-judge} evaluation \cite{zheng2023judging, li2025generation}, we therefore use a structured evaluator implemented within the \texttt{DeepEval} framework \cite{deepeval2025}.\footnote{\url{https://github.com/confident-ai/deepeval}}. A GPT-based judge model (GPT-4.1) is prompted to parse both predicted and gold Mermaid diagrams, normalize node labels, identify directed relations, and compute symbolic precision, recall, and F1-score for nodes and edges. The full evaluation prompts are provided in Appendix~\ref{appendix1}.

\subsection{Setup}

We evaluated five vision-language models integrated into our system: \texttt{GPT-4.1}, \texttt{GPT-4.1-mini}, \texttt{GPT-4o}, \texttt{GPT-4o-mini}, and \texttt{Gemini-2.5-Flash}. All models used the same structured instruction template and were tasked with generating Mermaid flowcharts for 200 input diagrams. Predicted Mermaid code was parsed automatically, normalized, and compared against gold-standard annotations.

\subsection{Symbolic Metrics}
These metrics quantify symbolic alignment and graph-topological similarity. We compute:

\noindent
\textbf{Entity extraction accuracy:} precision (P), recall (R), and F1-score between predicted and gold node labels.\\
\textbf{Relationship extraction accuracy:} P, R, and F1-score for directed edges.\\
\textbf{Semantic similarity:} Cosine similarity between the predicted and gold Mermaid code using SBERT embeddings \cite{reimers2019sentence}.

\subsection{High-Level Mermaid Structural Metrics}

To evaluate global diagram quality beyond exact node–edge matching, we use five complementary structural metrics assessed by GPT-4.1 (Appendix~\ref{appendix2}). Each metric captures a distinct aspect of diagram-level fidelity:

\noindent
\textbf{Structural Accuracy (0--1)}: high-level alignment of node–edge organization.\\
\textbf{Flow Accuracy (0--1)}: preservation of control-flow and execution paths.\\
\textbf{Syntax Validity (0--1)}: conformance to Mermaid syntax and renderability.\\
\textbf{Semantic Fidelity (0--1)}: retention of the workflow’s intended meaning.\\
\textbf{Completeness (0--1)}: inclusion of essential elements and relations from the gold diagram.


A \textbf{reconstructability override} is applied when the judge model determines that the predicted diagram could be transformed into the gold version through minor structure-preserving edits, granting full credit across metrics. Each metric is originally scored on its native bounded scale (\(0\!-\!5\), \(0\!-\!3\), or \(0\!-\!2\)) and subsequently normalized to the range \([0,1]\) for comparability. While these metrics are conceptually formalized, their implementation relies on GPT-4.1’s implicit reasoning rather than explicit numerical computation.

\begin{table*}[h]
\centering
\small
\begin{tabular}{lcccccccc}
\toprule
\textbf{Model} & \textbf{Entity P} & \textbf{Entity R} & \textbf{Entity F1} &
\textbf{Rel. P} & \textbf{Rel. R} & \textbf{Rel. F1} &
\textbf{Cosine Sim.} \\
\midrule
GPT-4.1 & 0.975 & 0.974 & 0.975 & 0.963 & 0.960 & 0.961 & 0.875 \\
GPT-4.1-mini & 0.986 & 0.986 & 0.986 & 0.980 & 0.978 & 0.979 & 0.858 \\
GPT-4o & 0.949 & 0.946 & 0.946 & 0.919 & 0.909 & 0.914 & 0.901 \\
GPT-4o-mini & 0.827 & 0.808 & 0.817 & 0.759 & 0.729 & 0.743 & 0.908 \\
Gemini-2.5-Flash & 0.986 & 0.987 & 0.986 & 0.976 & 0.976 & 0.976 & 0.863 \\
\bottomrule
\end{tabular}
\caption{Performance on a \textsc{FlowVQA} subset (200 diagrams).  
Precision (P), Recall (R), and F1-scores are reported for both entity and relationship extraction. Cosine similarity reflects semantic similarity between predicted and gold Mermaid code.}
\label{tab:evaluation}
\end{table*}

\section{Results and Discussion}

Tables~\ref{tab:evaluation} and~\ref{tab:highlevel} summarize performance on the \textsc{FlowVQA} subset. Table \ref{tab:evaluation} focuses on symbolic extraction quality (entities, relations, SBERT cosine similarity), while Table \ref{tab:highlevel} captures higher-level structural properties of the generated diagrams. For SBERT cosine similarity, we use a fine-tuned MiniLM \cite{wang2020minilm} model from Huggingface\footnote{\url{https://huggingface.co/sentence-transformers/all-MiniLM-L12-v2}}.

\paragraph{Symbolic extraction performance.}
All large models achieve strong entity extraction, with entity F1-scores above 0.94 for \texttt{GPT-4.1}, \texttt{GPT-4o}, and \texttt{Gemini-2.5-Flash}. Relationship extraction is consistently more difficult: relation F1 scores are typically 1-3 points lower than entity F1 scores, indicating that recovering the full edge structure is the primary source of error. \texttt{GPT-4.1-mini} and \texttt{Gemini-2.5-Flash} yield the strongest overall symbolic performance for both entity and relation F1-score. \texttt{GPT-4o} produces slightly lower relation F1-score, suggesting occasional omissions in branching structure while \texttt{GPT-4o-mini} is the weakest model overall. Cosine similarity scores are uniformly high, showing that label semantics are well preserved. Interestingly, \texttt{GPT-4o-mini} attains one of the highest cosine similarities despite poor structural F1-score, highlighting that embedding similarity alone cannot identify missing or misrouted transitions.

\paragraph{High-level structural behaviour.}
The high-level Mermaid metrics in Table~\ref{tab:highlevel} reveal how these symbolic errors translate into perceived diagram quality. For the strongest models, structural accuracy (SA), flow accuracy (FA), semantic fidelity (SF), and completeness (C) are all close to 1.0, indicating that the judge model regards the predicted diagrams as near-perfect reconstructions of the gold workflows. \texttt{Gemini-2.5-Flash} achieves the highest scores across these metrics, followed closely by \texttt{GPT-4.1-mini}. This suggests that on this dataset, the smaller \texttt{GPT-4.1-mini} is able to preserve the global process semantics almost as well as the largest models.

\texttt{GPT-4o} maintains perfect syntax validity (SV = 1.000) but shows slightly lower SA and FA (0.954), mirroring its lower relation F1-score. This pattern is consistent with a model that produces well-formed Mermaid code but sometimes simplifies or omits less salient branches. \texttt{GPT-4o-mini} again stands out: although its syntax validity is almost perfect (0.998), its SA, FA, SF, and C are substantially lower (around 0.83–0.86), confirming that it tends to generate syntactically correct but structurally incomplete diagrams.

\begin{table}[h]
\centering
\resizebox{\columnwidth}{!}{
\begin{tabular}{lccccc}
\toprule
Model &
SA &
FA &
SV &
SF &
C \\
\midrule
GPT-4.1           & 0.974 & 0.973 & 0.995 & 0.973 & 0.983 \\
GPT-4.1-mini      & 0.984 & 0.984 & 0.995 & 0.984 & 0.999 \\
GPT-4o            & 0.954 & 0.954 & 1.000 & 0.954 & 0.978 \\
GPT-4o-mini       & 0.831 & 0.830 & 0.998 & 0.829 & 0.861 \\
Gemini-2.5-Flash  & 0.988 & 0.988 & 0.998 & 0.988 & 0.999 \\
\bottomrule
\end{tabular}
}
\caption{
High-level Mermaid structural metrics on the same subset
(SA = structural accuracy; FA = flow accuracy; 
SV = syntax validity; SF = semantic fidelity; C = completeness).
}
\label{tab:highlevel}
\end{table}

\subsection{Discussion}

Taken together, the two metric families highlight complementary aspects of model behaviour. High entity and relation F1-score scores show that the best models recover the vast majority of symbolic content, while high SA, FA, SF, and C indicate that this content is arranged into globally coherent, semantically faithful workflows. The small but consistent gap between entity and relation performance confirms that edge reconstruction is more challenging than node labeling, and the discrepancy between cosine similarity and structural metrics for \texttt{GPT-4o-mini} illustrates that text-level similarity can mask important structural errors.

Practically, these results suggest that large multimodal models such as \texttt{GPT-4.1} and \texttt{Gemini-2.5-Flash} already support near-lossless flowchart-to-Mermaid translation on realistic diagrams, while smaller variants offer a controllable quality–latency trade-off. The complementary nature of symbolic and judge-based metrics is essential: symbolic metrics capture fine-grained correctness, whereas high-level structural metrics reveal whether a diagram remains interpretable as a coherent workflow.

\section{Conclusion and Future Work}

We introduced a lightweight web application that converts flowchart images into structured \texttt{Mermaid} code and supports mixed-initiative refinement through visual editing, code manipulation, and natural-language commands. Using a carefully engineered system prompt together with state-of-the-art VLMs, the system transforms static diagrams into editable, version-controllable representations. Future work includes expanding evaluation beyond the small \textsc{FlowVQA} subset, extending coverage to additional diagram types such as UML diagrams, exploring fine-tuned models specialized for diagram parsing, and enhancing interaction features such as intelligent autocompletion, collaborative editing, and cross-diagram consistency checking.

\section*{Limitations}

Flowchart2Mermaid converts flowchart images into editable Mermaid code through structured prompting and careful post-processing to ensure syntactic validity. However, as with all AI-based systems, the underlying vision–language models may occasionally hallucinate or misinterpret diagram elements. While the generated output typically provides a strong starting point, it may not exactly match the original diagram’s layout or semantics. The system is therefore intended as an assistive tool that complements human expertise, requiring user review and refinement to ensure correctness and completeness.

\section*{Acknowledgments}

We thank the maintainers of OpenAI and other VLM developers for access to their models and APIs. The authors acknowledge the use of generative AI tools (e.g., ChatGPT) for support in code implementation and text editing; all conceptual contributions and experimental results are original and have been independently verified. This research is supported by the Advanced Research and Engineering Centre (ARC) in Northern Ireland, funded by PwC and Invest NI. The views expressed are those of the authors and do not necessarily represent those of ARC or the funding organisations. 

\bibliography{anthology_0,custom}

\appendix

\section{Appendix 1}

\begin{tcolorbox}[enhanced,
    breakable,
    colback=white,
    colframe=black,
    width=\columnwidth,
    title={LLM-Based Mermaid Flowchart Evaluation Prompt},
    fonttitle=\bfseries,
    left=4pt,right=4pt,top=4pt,bottom=4pt]
\begin{lstlisting}[basicstyle=\ttfamily\scriptsize,breaklines=true]
You are an automated evaluator for Mermaid flowcharts.
You will be given two flowcharts: a prediction and a gold reference.

### Your tasks:
1. Parse both flowcharts.
   - Extract all nodes: ignore node IDs (like A, B, C...), shapes ([], (), {{}}, / /), and quote differences (' vs ` vs ").
   - Normalize whitespace and casing (so "Create a new user" == "create a new user").
   - Extract all relations (edges) as (source, target), ignoring arrow style (-->, -->|Yes|, etc.).

2. Compare prediction vs gold.
   - Count TP (correct matches), FP (predicted but not in gold), FN (in gold but missing in pred).
   - Do this separately for **nodes** and for **relations**.

3. Compute metrics:
   - Precision = TP / (TP + FP)   (0 if denominator is 0)
   - Recall    = TP / (TP + FN)   (0 if denominator is 0)
   - F1-score        = 2*P*R / (P+R)   (0 if both P and R are 0)

4. Output strictly:
   - **Only one line of CSV**
   - Exactly 6 numeric values, floats in decimal form
   - Order: nodes_precision,nodes_recall,nodes_F1-score,rels_precision,rels_recall,rels_F1-score
   - No headers, no labels, no text, no explanations, no extra lines

### Format example:
0.857,0.750,0.800,0.900,0.818,0.857

### Prediction
{pred}

### Gold
{gold}

\end{lstlisting}
\end{tcolorbox}
\label{appendix1}

\section{Appendix 2}

\begin{tcolorbox}[enhanced,
    breakable,
    colback=white,
    colframe=black,
    width=\columnwidth,
    title={LLM-Based High-Level Mermaid Flowchart Evaluation Prompt},
    fonttitle=\bfseries,
    left=4pt,right=4pt,top=4pt,bottom=4pt]
\begin{lstlisting}[basicstyle=\ttfamily\scriptsize,breaklines=true]
You are an automated evaluator for Mermaid flowcharts.
You will be given two flowcharts: a prediction and a gold reference.

Your job is to output ONE CSV line only, with 5 numbers (no text, no headers, no explanations).

Evaluate ONLY the following metrics:

1. structural_accuracy (0-5)
2. flow_accuracy        (0-5)
3. syntax_validity      (0-2)
4. semantic_fidelity    (0-5)
5. completeness         (0-3)


### VERY IMPORTANT RULE
If you believe the prediction can be **fully reconstructed into a valid Mermaid.js diagram** that accurately reflects the gold reference structure and meaning,
then **assign the maximum score to all five metrics**.

### Otherwise:
Score each metric according to the definitions:
- structural_accuracy: correctness of node and edge arrangement (0-5)
- flow_accuracy: correctness of overall process flow (0-5)
- syntax_validity: valid Mermaid syntax and proper graph structure (0-2)
- semantic_fidelity: how well the meaning matches the gold (0-5)
- completeness: how fully the prediction covers the gold (0-3)

### OUTPUT FORMAT (STRICT)
Output exactly ONE line:
structural_accuracy,flow_accuracy,syntax_validity,semantic_fidelity,completeness

### Prediction
{pred}

### Gold
{gold}

\end{lstlisting}
\end{tcolorbox}
\label{appendix2}

\end{document}